\documentclass{article}
\usepackage[T1]{fontenc}
\usepackage{arxiv}
\AtBeginDocument{\newgeometry{textheight=9in, textwidth=5.5in, top=1in,
  headheight=14pt, headsep=25pt, footskip=30pt}}
\usepackage{times}
\usepackage{natbib}
\setcitestyle{authoryear,round,citesep={;},aysep={,},yysep={;}}
\usepackage{amsmath,amsfonts,bm}

\def\eqref#1{equation~\ref{#1}}

\def\1{\bm{1}}

\def\vtheta{{\bm{\theta}}}

\DeclareMathAlphabet{\mathsfit}{\encodingdefault}{\sfdefault}{m}{sl}
\SetMathAlphabet{\mathsfit}{bold}{\encodingdefault}{\sfdefault}{bx}{n}

\usepackage{hyperref}
\usepackage{url}
\usepackage{amsmath,amssymb}
\usepackage{booktabs}
\usepackage{graphicx}
\usepackage{capt-of}
\usepackage{wrapfig}
\usepackage{algorithm}
\usepackage[noend]{algpseudocode}
\algrenewcommand{\algorithmiccomment}[1]{\hfill\makebox[0.5\linewidth][l]{$\triangleright$ #1}}
\newcommand{\CommentCont}[1]{\hfill\makebox[0.5\linewidth][l]{\phantom{$\triangleright$} #1}}
\usepackage{enumitem}
\usepackage{needspace}
\usepackage{xcolor}
\usepackage{colortbl}
\definecolor{heat1}{HTML}{F19E63}
\definecolor{heat2}{HTML}{F2BE85}
\definecolor{heat3}{HTML}{F4DBAE}
\definecolor{heat4}{HTML}{FBF4DA}
\newcommand{\hc}[1]{\cellcolor{heat#1}}
\usepackage[capitalise,noabbrev]{cleveref}
\AddToHook{env/table/begin}{\setlength{\abovecaptionskip}{0pt}\setlength{\belowcaptionskip}{6pt}}
\AddToHook{env/wraptable/begin}{\setlength{\abovecaptionskip}{0pt}\setlength{\belowcaptionskip}{6pt}}

\newcommand{\method}{\textbf{OmniHOI}}

\newcommand{\Lcal}{\mathcal{L}}

\newcommand{\vbeta}{{\bm{\beta}}}

\title{\method{}: Dexterous Hand-Object Interaction from Monocular Human Video}

\author{%
  Ting Mao$^{1}$, Yanming Shao$^{2}$, Ziheng Wang$^{3}$, Haoyu Liu$^{1}$, \\
  \bfseries Yiqun Wang$^{4}$, Xuanye Wu$^{4}$, and Yao Mu$^{3,4,\dagger}$ \\[6pt]
  \normalfont\small
  $^{1}$Zhejiang University \quad
  $^{2}$The University of Hong Kong \\
  \normalfont\small
  $^{3}$Shanghai Jiao Tong University \quad
  $^{4}$Shanghai AI Laboratory}
\date{}
\renewcommand{\undertitle}{}
\renewcommand{\headeright}{}
\renewcommand{\shorttitle}{OmniHOI: Dexterous Hand-Object Interaction from Monocular Human Video}
\hypersetup{pdftitle={OmniHOI: Dexterous Hand-Object Interaction from Monocular Human Video},
  pdfauthor={Ting Mao, Yanming Shao, Ziheng Wang, Haoyu Liu, Yiqun Wang, Xuanye Wu, Yao Mu}}
\begin{document}

\maketitle
{\renewcommand{\thefootnote}{$\dagger$}\footnotetext{Corresponding author.}}

\begin{center}
\includegraphics[width=\linewidth]{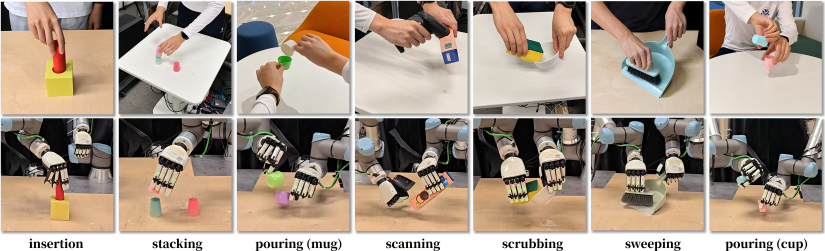}
\captionof{figure}{\textbf{From human video to robot execution.} Top: monocular RGB videos of human hand-object interaction. Bottom: the trajectories \method{} produces from them, executed on a pair of XHands.}
\label{fig:teaser}
\end{center}
\vspace{0.075in}

\begin{abstract}
Monocular videos of human manipulation provide abundant dexterous demonstrations, yet reconstructing hand-object interaction from a single view and transferring it to robot hands remain difficult, limiting their direct use for robot execution.
Prior methods either require task-specific RL training, limiting scalability, or assume clean motion-capture trajectories and thus cannot operate directly on video.
We present \method{}, a pipeline that turns an RGB video of hand-object interaction into an interaction-faithful trajectory on dexterous hands.
The key idea is to enforce physical consistency using the evidence available at each stage: image evidence during reconstruction, contact geometry during retargeting, and dynamics during physics-in-the-loop refinement. Each stage optimizes the corresponding representation directly, correcting errors before they propagate downstream or must be absorbed by a learned policy.
Across 150 motion-capture trajectories transferred to each of five dexterous hands with 6 to 22 DoF, we achieve 39--89\% success, compared with at most 31\% for prior transfer methods. On 60 monocular video clips, we achieve 53\% success, compared with 28\% for the best prior video-to-robot pipeline.
Its trajectories also execute on a real bimanual robot across diverse tasks.
The project page is available at \url{https://hydrofoooil.github.io/OmniHOI/}.
\end{abstract}

\section{Introduction}
\label{sec:intro}

Dexterous manipulation demonstrations must capture both hand motion and object interaction.
Such data are typically collected through teleoperation or wearable captures~\citep{wang2024dexcap,hoque2025egodex}, while human manipulation videos are far more abundant.
Recent monocular reconstruction models such as MoGe-3~\citep{kong2026moge3}, SAM 3D~\citep{chen2025sam3d}, and WiLoR~\citep{potamias2025wilor} can recover scene geometry, object meshes, and hand poses from a single view,
making these videos a promising source of robot demonstrations.
Turning such video into dexterous robot action requires more than recovering visually plausible motion.
Monocular depth ambiguity can confuse occlusion with contact, small geometric errors can turn contact into penetration or separation,
and a motion feasible for a human hand may become infeasible on a robot hand with different morphology and dynamics.
These errors must be resolved both when reconstructing the interaction and when transferring it to the robot.

Existing methods make different assumptions about the input and where errors are handled.
VideoManip~\citep{chen2026videomanip} and related methods~\citep{lum2025h2sim2robot,hsieh2025dexman,chen2026v2pmanip,mu2026deximit} reconstruct hand-object motion from video and then train a task-specific policy,
leaving the policy to compensate for reconstruction errors.
ManipTrans~\citep{li2025maniptrans} and SPIDER~\citep{pan2025spider} start from clean motion-capture trajectories and do not address noisy video reconstruction;
SPIDER also degrades when the reference motion is noisy.
Do as I Do~\citep{paliwal2026doasido} starts from video without training, but relates the hand and object mainly through centroid alignment and handles contact explicitly only in simulation.
Across these pipelines, geometric errors introduced early can survive retargeting and become harder to correct later.
A slight finger-object penetration in reconstruction, for example, can turn into a robot link passing through the object after retargeting, leaving the final optimizer with an invalid contact state.

We present \method{}, a pipeline that reconstructs hand-object interaction from video and transfers it to dexterous robot hands. During reconstruction, dense hand-scene correspondences recover the hand-object geometry,
followed by refinement for occlusion, contact, penetration, and finger crossing.
During retargeting, the robot hand is optimized to preserve demonstrated contact while removing morphology-induced penetration and crossings.
The resulting keyframes are then refined in simulation with CMA-ES optimizer~\citep{hansen2001cma} to improve object tracking, contact, and grasp stability.
Each stage corrects the errors exposed by its own evidence before they propagate to the next stage.

We evaluate \method{} on reconstruction, motion transfer, and end-to-end video-to-robot execution.
It achieves lower hand pose error than prior reconstruction methods while maintaining low hand-object intersection.
Across 150 motion-capture trajectories transferred to five dexterous hands with 6 to 22 DoF, we achieve 39--89\% success, compared with at most 31\% for prior transfer methods.
On 60 video clips, we achieve 53\% success, compared with 28\% for VideoManip and 25\% for Do as I Do, and execute the resulting trajectories on a real bimanual robot.
Our contributions are:
\begin{itemize}[leftmargin=*, itemsep=0pt, parsep=0pt]
\item A monocular video-to-dexterous-robot pipeline that maintains physical consistency across reconstruction, retargeting, and execution.
\item A hand-object reconstruction method that resolves occlusion, contact, penetration, and finger-crossing errors.
\item A contact- and dynamics-aware transfer method that produces task-effective motion across diverse dexterous hands.
\end{itemize}

\section{Related Work}
\label{sec:related}

\paragraph{Dexterous manipulation from human videos.}
Human videos have supplied manipulation
priors~\citep{nair2022r3m,ma2023vip,shaw2022videodex,singh2025hop,bahl2023vrb,agarwal2023functional,bharadhwaj2024track2act,wang2023mimicplay},
but these require robot data or target grippers, and precise dexterous
demonstrations still rely on gloves, AR glasses or
headsets~\citep{liu2025egozero,guzey2025aina}.
For functional grasping, FunGrasp~\citep{huang2025fungrasp} retargets human grasps
from a single RGB-D image to different dexterous hands and trains grasping policies
in simulation.
Closest to ours are pipelines that reconstruct the interaction from video and train
a policy on
it~\citep{qin2022dexmv}, like the systems discussed in \cref{sec:intro}.
Recent work also reconstructs hand-object motions from generated videos and grounds
them in simulation to train a multi-object, multi-trajectory
tracker~\citep{wu2026grounding}.

\paragraph{Monocular hand-object reconstruction.}
Hands are recovered robustly in the
wild~\citep{pavlakos2024hamer,zhang2025hawor}; the held object is
predicted by learned
models~\citep{hasson2019obman,wu2024mccho,chen2025hort} such as iHOI~\citep{ye2022ihoi}, lifted with large
pretrained models as in EasyHOI~\citep{liu2025easyhoi} and FollowMyHold~\citep{aytekin2025followmyhold}, or fit per
sequence~\citep{fan2024hold,ye2023diffhoi,ye2024ghop}. Fitting-based methods optimize
the two together under contact, collision and occlusion
constraints~\citep{hasson2021homan,cao2021rho,grady2021contactopt,jiang2020coherent,zhang2020phosa},
whereas robot-learning pipelines such as VideoManip and Do as I Do relate them only
through keypoint depth or centroids and leave the
contact--occlusion ambiguity open, which \method{} resolves with pixel-aligned
registration and per-finger occlusion-order, contact and crossing corrections.

\paragraph{Human-to-robot motion transfer.}
Kinematic retargeting, such as WujiHand Retargeting~\citep{wuji2026retargeting}, maps human keypoints to robot
joints~\citep{handa2020dexpilot,qin2023anyteleop,yin2025geort},
and interaction-preserving variants such as OmniRetarget~\citep{yang2025omniretarget} keep spatial or contact relations to the
object~\citep{ho2010spatial,lakshmipathy2025contact}, but both
preserve how the hand looks, not whether it holds the object.
Dynamics-aware methods track the reference in simulation through RL or
optimization~\citep{chen2024objdex,mandi2025dexmachina,liu2025dextrack,liu2024quasisim}
or training-free
sampling~\citep{williams2016mppi,howell2022predictive,liu2010samcon,ye2012contactsampling},
but many assume clean motion-capture references rather than video-derived trajectories.
\method{} resolves contact geometrically before simulation, then refines keyframe
offsets with CMA-ES in rolling horizons.

\section{Method}
\label{sec:method}

\begin{figure}[t]
\centering
\includegraphics[width=\linewidth]{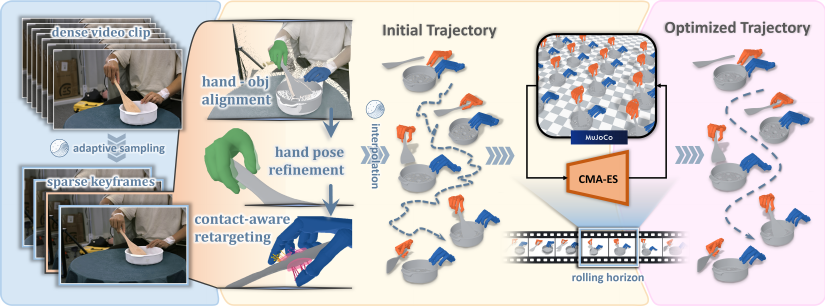}
\caption{\textbf{\method{} is built on several key designs.} Adaptive sampling reduces the video clip to keyframes \textbf{(left)}.
Hand pose in each keyframe is reconstructed, refined and retargeted with contact information preserved \textbf{(middle)}.
Interpolating between the keyframe configurations then gives an initial robot trajectory.
Physics-in-the-loop refinement then rolls the trajectory out in simulation and updates the keyframe configurations with CMA-ES, yielding a task-effective trajectory \textbf{(right)}.}
\label{fig:overview}
\end{figure}

As shown in \cref{fig:overview}, \method{} samples keyframes from the video according to hand motion,
reconstructs and refines the hand-object interaction in each keyframe (\cref{sec:method:recon}),
retargets the refined MANO hands~\citep{romero2017mano} to the robot hand under contact guidance,
and turns the keyframes into a continuous, task-effective trajectory in simulation (\cref{sec:method:exec}).

\par\penalty10000
\ifdim\dimexpr\pagegoal-\pagetotal\relax>0pt
  \ifdim\dimexpr\pagegoal-\pagetotal\relax<19\baselineskip\newpage\fi\fi
\subsection{Keyframe reconstruction}
\label{sec:method:recon}

\begin{wrapfigure}[15]{r}{2.55in}
\vspace{-8pt}
\centering
\includegraphics{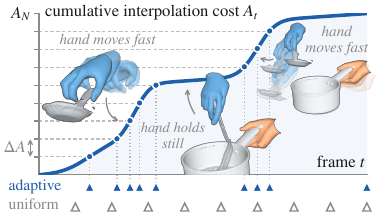}
\caption{\textbf{Keyframes are equal increments of $A_t$.} Equal steps $\Delta \mathrm{A}$ are unequal in time: keyframes crowd where the hand moves fast, unlike uniform sampling of the same budget.}
\label{fig:keyframes}
\end{wrapfigure}
We use SAM 3~\citep{carion2025sam3} to segment hands, objects and table, and use MoGe-3 to estimate camera intrinsics and each frame's depth as point map.
With a static camera, every point map is aligned to the first frame on table pixels, whose fitted plane later defines the simulation world frame.

\paragraph{Object-centric hand placement.}
WiLoR reconstructs the hand mesh up to unknown depth and scale.
As mesh and point map are \emph{pixel-aligned}, rasterizing the visible MANO surface pairs each hand-mask pixel with a vertex $m_i$ and a scene point $x_i$, over which we fit a similarity $(s_h, \mathbf{R}_h, \mathbf{t}_h)$.
A first, unweighted fit compares the map with the MANO hand's real size; as monocular metric scale is unreliable, its median over all hands and frames fixes one unit for the clip.
In this unit, SAM 3D reconstructs each object once from its most visible frame, and FoundationPose~\citep{wen2024foundationpose} registers it per frame from its mask.
The final fit weights each pair by its distance $d_i$ to the object, $w_i=\exp(-\kappa\,d_i/\ell)$ with $\ell$ the index distal phalanx length, as monocular depth is most reliable near the object.

\paragraph{Adaptive keyframe sampling.}
Refining every frame is expensive, and at frame scale the coarse poses are dominated by jitter rather than by real motion,
so we refine only $r \cdot N$ of the $N$ frames and recover the rest by interpolation.
We first accumulate the second-order motion of each hand into
\begin{equation}
A_t=\sum_{u\leq t}\bigg(\omega_{\mathrm{trans}}\big\|\delta^2\mathbf{t}_u\big\|
+\omega_{\mathrm{rot}}\big\|\delta^2\mathbf{R}_u\big\|_F
+\omega_{\mathrm{joint}}\sum_{j=1}^{n_{\mathrm{joints}}}\big\|\delta^2\vtheta_{j,u}\big\|\bigg)^{1/2},
\label{eq:motionsignal}
\end{equation}
where $\mathbf{t}_u$, $\mathbf{R}_u$ and $\vtheta_{j,u}$ denote wrist translation, wrist orientation and joint angles at frame $u$,
and $\delta^2(\cdot)$ is the central second-order difference of the smoothed trajectory.
Keyframes are placed at equal increments of $A_t$, so they crowd at reversals and transients and are sparse on steady stretches (\cref{fig:keyframes}; details in \cref{app:keyframes}).

\paragraph{Alternating image-contact refinement.}
The roughly placed hand may still have incorrect occlusion, lost contact or penetration.
The two error types are corrected by different evidence and different variables,
so we alternate two stages:
\emph{image alignment} adjusts the articulation $\vtheta$ and hand shape $\vbeta$,
\emph{contact-penetration optimization} adjusts $\vtheta$ together with $\mathbf{T}_h = (\mathbf{R}_h, \mathbf{t}_h)$.

\begin{wrapfigure}[21]{r}{2.75in}
\vspace{-8pt}
\centering
\includegraphics{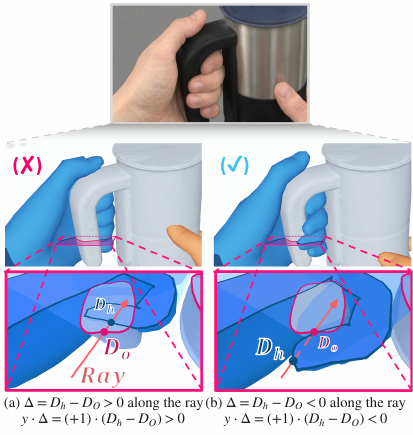}
\caption{\textbf{Occlusion-order loss.} Input image and handle close-ups before (a) and after (b) refinement.
A ray through an overlap pixel meets the hand at depth $D_h$ and the object at depth $D_O$;
the order is violated in (a) and correct in (b), where the fingers no longer intersect the handle.}
\label{fig:occlusion}
\end{wrapfigure}

\emph{Image-alignment optimization.}
We render the hand and object depths $D_h$, $D_O$ and the soft hand silhouette $S_h$ with a differentiable rasterizer~\citep{ravi2020pytorch3d}.
For a pixel $p$ in the hand-object overlap $\Omega$,
let $\Delta(p)=D_h(p)-D_O(p)$,
and $y(p)=+1$ when the hand is visible and $y(p)=-1$ when the object is visible;
correct ordering satisfies $y(p)\Delta(p)<0$ (\cref{fig:occlusion}).
We penalize violations by
\begin{equation}
\Lcal_{\mathrm{ord}}=\frac{\sum_{p\in\Omega}w(p) \cdot \operatorname{sigmoid}\big(y(p)\Delta(p)/s_{\mathrm{o}}\big)}{\sum_{p\in\Omega}w(p)},\qquad w(p)=\operatorname{sg}[S_h(p)],
\label{eq:ord}
\end{equation}
with $s_{\mathrm{o}}$ a depth scale.
The stop-gradient $\operatorname{sg}[\cdot]$ freezes $w$ during back-propagation,
so the hand cannot lower the loss by shrinking its overlap instead of correcting the order.
The term is evaluated per finger.

\par\penalty10000
\ifdim\dimexpr\pagegoal-\pagetotal\relax>0pt
  \ifdim\dimexpr\pagegoal-\pagetotal\relax<27\baselineskip\newpage\fi\fi
\emph{Contact-penetration optimization.}
This stage attracts each finger's palmar samples $P_f$ to the object surface while repelling its full-surface samples $Q_f$ from the interior.
With $\phi$ the object's signed distance (positive outside) and $\langle\cdot\rangle_P$ an area-weighted mean over $P$,

\begin{equation}
\begin{gathered}
E^{c}_f=\Big\langle\tfrac{d}{\sigma_c}\log\big(1+\tfrac{d}{\sigma_c}\big)\Big\rangle_{P_f},\\
E^{p}_f=\Big\langle\big(\operatorname{ReLU}(-\phi-d_0)/\sigma_p\big)^2\Big\rangle_{Q_f},\\
\Lcal_{\mathrm{grasp}}=\frac15\sum_{f=1}^{5}\big(\lambda_cE^{c}_f+\lambda_p\,e^{a_f}E^{p}_f\big),
\end{gathered}
\label{eq:grasp}
\end{equation}

where $d=\operatorname{ReLU}(\phi)$, $d_0$ is a penetration tolerance and $a_f$ the object-surface area enclosed by finger $f$,
an exponential weighting that we found necessary to keep deeply wrapped fingers from sinking into the object.
\Cref{fig:occlusion} shows the resulting change in finger-object occlusion.

\emph{Finger-crossing repair.} Signed distance saturates on thin objects:
a finger that has \emph{passed through} a plate reads a constant depth and slides inside with no gradient toward the exit (\cref{fig:retarget}).
To handle fingers that pass entirely through thin objects, we detect surface crossings using parity along each finger's keypoint polyline.
We then solve for the minimum pose change that removes the crossing residuals, using a damped linearized update with line search.
Full procedure is given in \cref{alg:crossing}.
The two stages alternate for a fixed number of cycles; repair runs after every cycle. Details are in \cref{app:recon}.

\subsection{Interaction-faithful human-to-robot transfer}
\label{sec:method:exec}

Keyframe reconstruction provides hand poses aligned with the object.
To turn these poses into executable robot motion, we first retarget them
to the robot hand while preserving contact information,
then we interpolate the retargeted keyframes at their source timestamps
and refine them in simulation to improve grasp stability and object
tracking.

\begin{figure}[t]
\centering
\includegraphics[width=\linewidth]{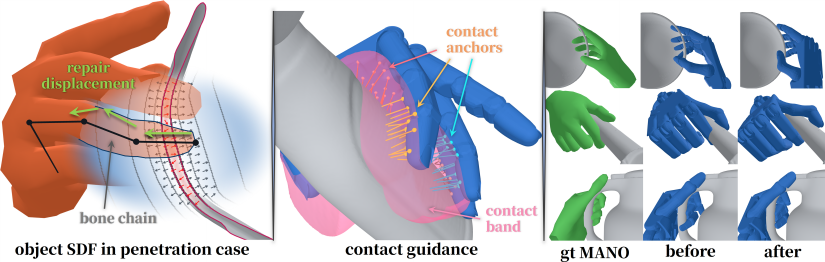}
\caption{\textbf{Contact-aware retargeting.}
\textbf{Left:} Arrows denote the gradient directions provided by the SDF. Gradients acting on the same finger link point in opposite directions and cancel each other out,
but the repair makes the finger be effectively pushed out of the object along the bone chain.
\textbf{Middle:} contact guidance: the anchors are sampled on the robot surface within the band around the
contact region $R_f$, which is required to reach. \textbf{Right:} comparisons between ground-truth MANO hand, the retargeting solver's
output, and the result after repair. With repair, the robot finger remains penetration-free even when the noisy ground-truth MANO annotation already penetrates the object.}
\label{fig:retarget}
\end{figure}

\paragraph{Contact-aware retargeting.}
We initialize the robot configuration $q^{\mathrm{init}}$ with the
keypoint-vector solver of WujiHand Retargeting.
This preserves the overall finger arrangement, but morphological
differences cause contact loss and penetration. We detect hand-object contacts by proximity,
assign each contacting finger a contact region $R_f$ on the object and a set of surface anchors,
and require its anchors to lie in a narrow band around $R_f$ (\cref{fig:retarget}) and prevent penetration at the same time.
Similar to \cref{sec:method:recon}, starting from the kinematic retargeting solution, we iteratively refine wrist and joint configurations.
Crossing links are repaired first; otherwise, the update jointly removes penetration/enclosure and pulls contact anchors toward their demonstrated contact regions.
We take the smallest pose change that lexicographically reduces the number of crossings and then the residual. See \cref{alg:retarget} for pseudocode.

\paragraph{Physics-in-the-loop trajectory refinement.}
We interpolate the retargeted keyframes $\{q_k\}$ at their source timestamps $\{t_k\}$ into a 60\,Hz trajectory by shape-preserving cubic interpolation \citep[PCHIP;][]{fritsch1980monotone},
then we replay the trajectory in MuJoCo~\citep{todorov2012mujoco} and search for corrections.
The variables for optimization are a per-keyframe offset $\Delta q_k$, interpolated the same way and added to the base trajectory.

Contact simulation provides no usable gradients,
so we optimize the offsets with CMA-ES, which maintains a multivariate Gaussian over them:
each iteration draws samples, scores them by rollouts, and shifts and reshapes it toward the lowest-cost ones,
so that it gradually moves toward and contracts around a minimum.
The search runs in a rolling horizon: at keyframe $k$ the simulator resumes from the state saved at $k-1$ and rolls out to $\max(t_{k+1},t_k+1\,\mathrm{s})$.
Each candidate is scored by
\begin{equation}
\begin{aligned}
\Lcal ={}& \lambda_{\mathrm{pos}}E_{\mathrm{pos}} + \lambda_{\mathrm{rot}}E_{\mathrm{rot}} + \lambda_{\mathrm{contact}}E_{\mathrm{contact}} + \lambda_{\mathrm{opp}}E_{\mathrm{opp}} \\
&+ \lambda_{\mathrm{pen}}E_{\mathrm{pen}} + \lambda_{\mathrm{prior}}\|\Delta q_k\|^2 + \lambda_{\mathrm{term}}E_{\mathrm{term}}.
\end{aligned}
\label{eq:physics_refinement}
\end{equation}
$E_{\mathrm{pos}}$ and $E_{\mathrm{rot}}$ measure the error between simulated object pose and its reconstructed trajectory,
and $E_{\mathrm{pen}}$ penalizes finger-object penetration.
$E_{\mathrm{contact}}$ is a penalty on the gap between object and hand links that the retargeted result marks as touching,
$E_{\mathrm{opp}}$ penalizes the shortfall of the best antagonistic contact pair below a target force,
and $E_{\mathrm{term}}$ re-applies the pose and contact penalties at the final step of the window,
penalizing grasps that look adequate mid-window but have already begun to fail by its end --- once $\Delta q_k$ is frozen, no later keyframe can undo it.

The best offset is then frozen, the simulation advances to $t_k$, and the search at $k+1$ warm-starts from $\Delta q_k$.

\section{Experiments}
\label{sec:exp}

\subsection{Hand-object reconstruction}
\label{sec:exp:recon}

\begin{figure}[t]
\centering
\includegraphics[width=\linewidth]{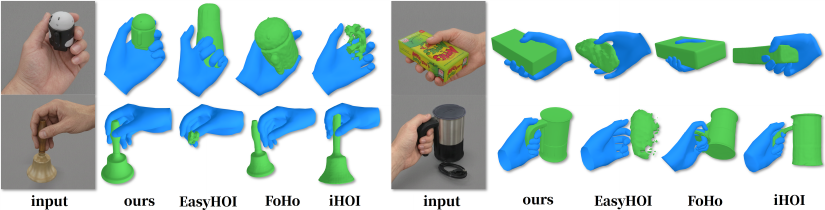}
\caption{\textbf{Qualitative hand-object reconstruction} from a single image. FoHo: FollowMyHold.}
\label{fig:recon_qual}
\end{figure}

We follow the setup and criteria from prior work, testing our method on three public datasets:
ARCTIC~\citep{fan2023arctic}, DexYCB~\citep{chao2021dexycb} and OakInk2~\citep{zhan2024oakink2}, all with 3D hand-object pose and shape annotations.
ARCTIC includes humans operating 11 articulated objects, OakInk2 features manipulation tasks with 75 diverse objects, and DexYCB presents grasping of 20 distinct YCB-Video~\citep{xiang2018posecnn} objects.
Since prior monocular hand-object reconstruction methods handle only one hand (ours handles both), we compare on 500 randomly sampled images per dataset, each showing one hand in contact with the object.

\begin{table}[h]
\centering
\caption{\textbf{Reconstruction results.} F5/F10: object F-score at a 5/10\,mm surface-distance threshold; CD: object Chamfer distance; MPVPE/MPJPE: mean per-vertex/per-joint error of the hand with dataset ground truth after Procrustes alignment; IV: hand-object intersection volume. Deeper color is better.}
\label{tab:recon}
\scriptsize
\setlength{\tabcolsep}{3.2pt}
\begin{tabular}{l cccccc cccccc}
\toprule
 & \multicolumn{6}{c}{Overall} & \multicolumn{6}{c}{ARCTIC} \\
\cmidrule(lr){2-7} \cmidrule(lr){8-13}
Method & F5 & F10 & CD & MPVPE & MPJPE & IV & F5 & F10 & CD & MPVPE & MPJPE & IV \\
 & $\uparrow$ & $\uparrow$ & (cm$^2$) $\downarrow$ & (mm) $\downarrow$ & (mm) $\downarrow$ & (cm$^3$) $\downarrow$ & $\uparrow$ & $\uparrow$ & (cm$^2$) $\downarrow$ & (mm) $\downarrow$ & (mm) $\downarrow$ & (cm$^3$) $\downarrow$ \\
\midrule
EasyHOI                                      & \hc{2}0.2187 & \hc{2}0.3890 & \hc{2}1.7110 & \hc{4}14.63 & \hc{4}14.02 & \hc{4}23.66 & \hc{3}0.1720 & \hc{3}0.3067 & \hc{3}2.6391 & \hc{4}16.20 & \hc{4}15.43 & \hc{3}27.19 \\
FollowMyHold                                 & \hc{4}0.2007 & \hc{4}0.3602 & \hc{4}1.7731 & \hc{2}8.92 & \hc{2}8.48 & \hc{3}19.73 & \hc{4}0.1698 & \hc{4}0.3035 & \hc{2}2.6325 & \hc{2}10.69 & \hc{2}10.10 & \hc{4}28.30 \\
iHOI                                         & \hc{3}0.2158 & \hc{3}0.3881 & \hc{3}1.7384 & \hc{3}11.41 & \hc{3}10.88 & \hc{2}2.90 & \hc{2}0.1741 & \hc{2}0.3137 & \hc{4}2.6672 & \hc{3}12.79 & \hc{3}12.07 & \hc{1}2.35 \\
\method{} (ours)                             & \hc{1}0.3878 & \hc{1}0.5831 & \hc{1}0.9574 & \hc{1}6.05 & \hc{1}5.71 & \hc{1}2.75 & \hc{1}0.2823 & \hc{1}0.4568 & \hc{1}1.6048 & \hc{1}5.36 & \hc{1}4.99 & \hc{2}2.64 \\
\midrule
 & \multicolumn{6}{c}{DexYCB} & \multicolumn{6}{c}{OakInk2} \\
\cmidrule(lr){2-7} \cmidrule(lr){8-13}
 & F5 & F10 & CD & MPVPE & MPJPE & IV & F5 & F10 & CD & MPVPE & MPJPE & IV \\
 & $\uparrow$ & $\uparrow$ & (cm$^2$) $\downarrow$ & (mm) $\downarrow$ & (mm) $\downarrow$ & (cm$^3$) $\downarrow$ & $\uparrow$ & $\uparrow$ & (cm$^2$) $\downarrow$ & (mm) $\downarrow$ & (mm) $\downarrow$ & (cm$^3$) $\downarrow$ \\
\midrule
EasyHOI                                      & \hc{3}0.2592 & \hc{3}0.4672 & \hc{2}0.7856 & \hc{4}12.93 & \hc{4}12.47 & \hc{4}28.57 & \hc{2}0.2300 & \hc{2}0.4013 & \hc{3}1.6193 & \hc{4}14.62 & \hc{4}14.02 & \hc{3}14.31 \\
FollowMyHold                                 & \hc{4}0.2127 & \hc{4}0.3890 & \hc{4}1.0262 & \hc{2}5.79 & \hc{2}5.60 & \hc{3}14.95 & \hc{3}0.2239 & \hc{3}0.3952 & \hc{2}1.5709 & \hc{2}10.21 & \hc{2}9.67 & \hc{4}14.91 \\
iHOI                                         & \hc{2}0.2611 & \hc{2}0.4708 & \hc{3}0.8182 & \hc{3}10.50 & \hc{3}10.13 & \hc{1}3.31 & \hc{4}0.2159 & \hc{4}0.3865 & \hc{4}1.6404 & \hc{3}10.77 & \hc{3}10.29 & \hc{2}3.10 \\
\method{} (ours)                             & \hc{1}0.4875 & \hc{1}0.7159 & \hc{1}0.3436 & \hc{1}5.03 & \hc{1}4.88 & \hc{2}4.44 & \hc{1}0.4040 & \hc{1}0.5883 & \hc{1}0.8594 & \hc{1}7.94 & \hc{1}7.43 & \hc{1}1.09 \\
\bottomrule
\end{tabular}
\end{table}

Our pipeline outperforms all baselines (iHOI, FollowMyHold and EasyHOI) on object reconstruction (F5, F10 and CD) and attains substantially lower hand pose errors (MPVPE and MPJPE).
Its intersection volume is on par with iHOI (2.75 vs.\ 2.90\,cm$^3$ overall) and far below those of FollowMyHold and EasyHOI.
\cref{fig:recon_qual} compares the reconstructions qualitatively.

\subsection{Human-to-robot motion transfer}
\label{sec:exp:transfer}

\begin{wrapfigure}{R}{2.75in}
\centering
\includegraphics[width=\linewidth]{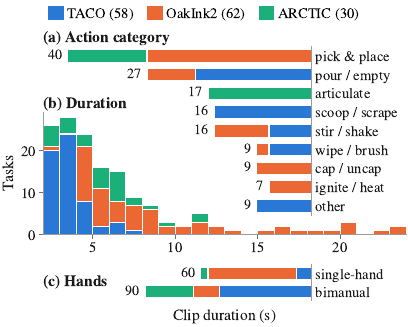}
\caption{\textbf{The 150 transfer tasks} by source dataset. (a)~Action category, where ``other'' is cutting and measuring. (b)~Duration of the replayed in-hand clips. (c)~Single-hand versus bimanual tasks.}
\label{fig:transfer_tasks}
\end{wrapfigure}

We compare our method with existing motion transfer approaches to evaluate its ability to convert human hand motion into functionally equivalent dexterous hand action given clean motion-capture data.
We select 150 trajectories covering diverse manipulation types, 58 from TACO~\citep{liu2024taco}, 62 from OakInk2 and 30 from ARCTIC,
of which 60 are single-hand and 90 bimanual (\cref{fig:transfer_tasks}),
retarget them to five dexterous hands using our method and the baselines (ManipTrans, SPIDER and OmniRetarget),
and evaluate manipulation precision, task compliance and transfer efficiency.

\paragraph{Setup.} Output trajectories are replayed open-loop in an identical MuJoCo scene. Each scene contains the ground-truth object as a free rigid body with annotated mass, the table, and the remaining static objects. Each hand uses its official model with its joint damping and torque limits; all other simulation settings, including actuator gains, are shared across methods (\cref{app:sim}).

\paragraph{Metrics.} For each task we evaluate how far the trajectory of each manipulated object in the replay deviates from its ground-truth trajectory,
since the goal is to move the object correctly rather than to precisely imitate the human hand.
For each object, we rigidly align the replayed poses $(\mathbf{p}_t,\mathbf{R}_t)$ to the ground truth $(\hat{\mathbf{p}}_t,\hat{\mathbf{R}}_t)$ and take the absolute trajectory error (ATE) ${e_{\mathrm{pos}}=\operatorname{RMS}_t\|\mathbf{p}_t-\hat{\mathbf{p}}_t\|}$ and ${e_{\mathrm{rot}}=\operatorname{RMS}_t\angle(\mathbf{R}_t,\hat{\mathbf{R}}_t)}$, with $\angle$ the geodesic angle.
A task is successful only if ATE stays below 10\,cm and 0.5\,rad, and in bimanual tasks both objects must pass.
Reported errors are per-task worst-object values, aggregated as the median over the tasks of each hand.

\begin{table}[t]
\centering
\caption{Human-to-robot motion transfer on 150 trajectories per hand across different robot hands. ManipTrans does not support Wuji Hand and SharpaWave. Time is the mean wall-clock minutes to process a trajectory over all hands a method was run on, measured on RTX 4090. It covers per-task RL training and rollout for ManipTrans (excluding imitator pre-training), whole pipeline for SPIDER, and human-to-robot motion transfer for \method{}. Deeper color is better.}
\label{tab:transfer}
\scriptsize
\setlength{\tabcolsep}{2.0pt}
\begin{tabular}{@{}l ccc ccc ccc ccc ccc c@{}}
\toprule
 & \multicolumn{3}{c}{Inspire Hand (6DoF)} & \multicolumn{3}{c}{XHand (12DoF)} & \multicolumn{3}{c}{Wuji Hand (20DoF)} & \multicolumn{3}{c}{SharpaWave (22DoF)} & \multicolumn{3}{c}{Shadow Hand (20DoF)} & \\
\cmidrule(lr){2-4} \cmidrule(lr){5-7} \cmidrule(lr){8-10} \cmidrule(lr){11-13} \cmidrule(lr){14-16}
Method & Succ. & Trans. & Rot. & Succ. & Trans. & Rot. & Succ. & Trans. & Rot. & Succ. & Trans. & Rot. & Succ. & Trans. & Rot. & Time \\
 & (\%) $\uparrow$ & (cm) $\downarrow$ & ($^\circ$) $\downarrow$ & (\%) $\uparrow$ & (cm) $\downarrow$ & ($^\circ$) $\downarrow$ & (\%) $\uparrow$ & (cm) $\downarrow$ & ($^\circ$) $\downarrow$ & (\%) $\uparrow$ & (cm) $\downarrow$ & ($^\circ$) $\downarrow$ & (\%) $\uparrow$ & (cm) $\downarrow$ & ($^\circ$) $\downarrow$ & (min) $\downarrow$ \\
\midrule
ManipTrans            & \hc{3}24 & \hc{4}9.4 & \hc{3}37 & \hc{4}22 & \hc{4}9.7 & \hc{2}33 & -- & -- & -- & -- & -- & -- & \hc{2}26 & \hc{4}10.7 & \hc{2}34 & \hc{4}385 \\
SPIDER                   & \hc{2}26 & \hc{1}6.6 & \hc{4}43 & \hc{3}26 & \hc{3}6.2 & \hc{4}43 & \hc{3}20 & \hc{3}8.3 & \hc{2}43 & \hc{3}19 & \hc{3}8.1 & \hc{3}54 & \hc{4}15 & \hc{3}8.5 & \hc{3}45 & \hc{3}37 \\
OmniRetarget      & \hc{3}24 & \hc{3}8.4 & \hc{2}33 & \hc{2}31 & \hc{2}3.8 & \hc{3}35 & \hc{2}27 & \hc{2}6.4 & \hc{2}43 & \hc{2}20 & \hc{2}7.3 & \hc{2}46 & \hc{3}16 & \hc{2}8.1 & \hc{4}50 & \hc{1}0.3 \\
\method{} (ours)                               & \hc{1}39 & \hc{2}6.7 & \hc{1}26 & \hc{1}89 & \hc{1}1.8 & \hc{1}12 & \hc{1}86 & \hc{1}1.6 & \hc{1}11 & \hc{1}89 & \hc{1}1.5 & \hc{1}10 & \hc{1}85 & \hc{1}2.4 & \hc{1}14 & \hc{2}19 \\
\bottomrule
\end{tabular}
\end{table}

\paragraph{Results.} As shown in \cref{tab:transfer}, our method achieves the highest success rate on every hand,
85--89\% on the four higher-DoF hands where the baselines reach at most 31\%,
and the lowest object errors except for translation on the Inspire Hand,
indicating that the motions it produces have reliable interaction fidelity.
The lead holds on all five hands, from 6 to 22 DoF, demonstrating strong cross-embodiment generalization.
Success on the 6-DoF Inspire Hand (39\%) is nevertheless clearly lower, which we attribute to the limited dexterity of its six motors.
In transfer efficiency, \method{} takes 19\,min per trajectory, second only to the purely kinematic OmniRetarget (0.3\,min) and below SPIDER (37\,min) and ManipTrans (385\,min), while succeeding far more often than all three.

\subsection{End-to-end video-to-robot transfer}
\label{sec:exp:e2e}

From the 150 videos collected in the previous section,
we select the 60 videos with the least hand-object occlusion and the largest hand-object pixel coverage in the frames to evaluate the quality of the outputs produced by the full video-to-action pipeline.
They comprise 51 TACO and 9 OakInk2 clips; none comes from ARCTIC, since we avoid articulated objects for reconstruction.
We use XHand as robot embodiment for evaluation.

\begin{table}[t]
\centering
\caption{\textbf{End-to-end video-to-robot transfer} on the 60 clips, under the ATE criterion of \cref{sec:exp:transfer}. Succ.Prog.: Progress fraction of the longest prefix that passes the success criterion (mean).}
\label{tab:e2e}
\scriptsize
\setlength{\tabcolsep}{3pt}
\begin{tabular}{l ccccc}
\toprule
Method & Succ. & Succ.Prog. & Trans. & Rot.  \\
 & (\%) $\uparrow$ & (\%) $\uparrow$ & (cm) $\downarrow$ & ($^\circ$) $\downarrow$  \\
\midrule
VideoManip & 28 & 49 & 8.6 & 31  \\
Do as I Do & 25 & 48 & 9.1 & 31  \\
\method{} (ours) & \textbf{53} & \textbf{71} & \textbf{5.7} & \textbf{26}  \\
\bottomrule
\end{tabular}
\end{table}
\paragraph{Failure modes.}
Both baselines estimate the camera with MoGe~\citep{wang2025moge,wang2025moge2}. On TACO it reads the 24$^\circ$ telephoto lens as 42\nobreakdash--52$^\circ$ and places the whole scene at about half its true distance,
so upstream image-based geometry estimation is their largest bottleneck;
on OakInk2, whose camera it estimates well, the hand pushes the object aside instead of grasping it, a second bottleneck independent of depth (\cref{app:e2e_failures}).
\method{} fails on 28 clips, 14 in perception, 13 in execution and 1 in motion planning,
most often because the object is reconstructed at the wrong size (12) or rotates in the hand (7).

\subsection{Real-world deployment}
\label{sec:exp:real}

\begin{figure}[h]
\centering
\includegraphics[width=\linewidth]{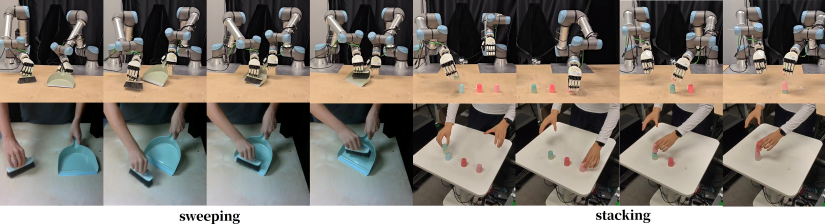}
\caption{\textbf{Real-world deployment.} Two representative tasks, sweeping and stacking, in temporal order from left to right. Top: the robot executing the trajectory. Bottom: input human video.}
\label{fig:realworld}
\end{figure}

To demonstrate quality, we execute seven representative self-recorded trajectories on two XHands mounted on UR5e arms,
each posing challenges such as bimanual coordination, precise insertion and long-horizon grasp-and-place.
We first align the hand-object trajectories produced by our pipeline with the robot workspace in simulation.
Since our pipeline operates on floating hands throughout, we then solve IK in simulation to obtain the corresponding arm motions.
Finally, we reproduce the simulated scene in the real world and execute the trajectories (\cref{fig:realworld});
further film strips are in \cref{app:realworld}.

\subsection{Ablations}
\label{sec:exp:ablation}

We ablate reconstruction and transfer separately, each paired over the same frames or tasks;
adaptive keyframe sampling is evaluated in \cref{app:keyframes}.

\par\penalty10000
\ifdim\dimexpr\pagegoal-\pagetotal\relax>0pt
  \ifdim\dimexpr\pagegoal-\pagetotal\relax<14\baselineskip\newpage\fi\fi
\begin{wraptable}{r}{2.75in}
\centering
\caption{Reconstruction ablations on the 1500 frames of \cref{tab:recon}: starting from the hand placed as in \cref{sec:method:recon}, the refinement steps of \cref{sec:method:recon} (image alignment with $\Lcal_{\mathrm{ord}}$, contact-penetration with $\Lcal_{\mathrm{grasp}}$, finger-crossing repair) are added one at a time.}
\label{tab:ablation_recon}
\scriptsize
\setlength{\tabcolsep}{4pt}
\begin{tabular}{l ccc}
\toprule
Variant & PA-MPVPE & PA-MPJPE & IV \\
 & (mm) $\downarrow$ & (mm) $\downarrow$ & (cm$^3$) $\downarrow$ \\
\midrule
Placement only & \textbf{5.37} & \textbf{5.03} & 4.27 \\
\quad + image alignment & 5.84 & 5.50 & 3.61 \\
\quad + contact-penetration & 5.81 & 5.48 & 3.03 \\
\quad + finger-crossing repair & 6.05 & 5.71 & \textbf{2.75} \\
\bottomrule
\end{tabular}
\end{wraptable}
\paragraph{Reconstruction components.}
Adding the refinement steps of \cref{sec:method:recon} one at a time to the placed WiLoR hand (\cref{tab:ablation_recon}) lowers the intersection volume at every step,
by about a third overall (from 4.27 to 2.75\,cm$^3$), while PA-MPVPE and PA-MPJPE rise by only 0.67\,mm and remain below every baseline in \cref{tab:recon}.
We accept this trade because retargeting inherits penetration: a robot hand that starts inside the object cannot separate from it in simulation.

\paragraph{Transfer components.}
We switch contact refinement and physics-in-the-loop refinement on and off on the protocol of \cref{sec:exp:transfer} (\cref{tab:ablation_transfer});
without physics-in-the-loop refinement, the interpolated keyframes are replayed open-loop.
Physics-in-the-loop refinement makes the motion executable, raising success from 9--29\% to 85--89\% on the four higher-DoF hands and from 21--25\% to 36--39\% on the Inspire Hand.
Replayed open-loop, contact refinement lowers success on four hands, mostly by removing false successes:
of the 51 tasks it turns into failures, 38 have fingers passing fully through the object in the $q^{\mathrm{init}}$ replay (5 after contact refinement),
so the interlocked object follows the hand without being held.
After physics-in-the-loop refinement it roughly halves the median maximum penetration depth, from 9.17 to 4.73\,mm,
while success rises by 2--3 points on three hands, is unchanged on the Shadow Hand and drops by 2 on the Wuji Hand (\cref{app:transfer_ablation}).

\begin{table}[h]
\centering
\caption{Transfer ablations: the two refinement steps of \cref{sec:method:exec} switched on and off, on the protocol of \cref{tab:transfer} (150 tasks per hand, from ground-truth MANO, ATE criterion). Deeper color is better.}
\label{tab:ablation_transfer}
\scriptsize
\setlength{\tabcolsep}{2.2pt}
\begin{tabular}{l ccc ccc ccc ccc ccc}
\toprule
 & \multicolumn{3}{c}{Inspire Hand (6DoF)} & \multicolumn{3}{c}{XHand (12DoF)} & \multicolumn{3}{c}{Wuji Hand (20DoF)} & \multicolumn{3}{c}{SharpaWave (22DoF)} & \multicolumn{3}{c}{Shadow Hand (20DoF)} \\
\cmidrule(lr){2-4} \cmidrule(lr){5-7} \cmidrule(lr){8-10} \cmidrule(lr){11-13} \cmidrule(lr){14-16}
Variant & Succ. & Trans. & Rot. & Succ. & Trans. & Rot. & Succ. & Trans. & Rot. & Succ. & Trans. & Rot. & Succ. & Trans. & Rot. \\
 & (\%) $\uparrow$ & (cm) $\downarrow$ & ($^\circ$) $\downarrow$ & (\%) $\uparrow$ & (cm) $\downarrow$ & ($^\circ$) $\downarrow$ & (\%) $\uparrow$ & (cm) $\downarrow$ & ($^\circ$) $\downarrow$ & (\%) $\uparrow$ & (cm) $\downarrow$ & ($^\circ$) $\downarrow$ & (\%) $\uparrow$ & (cm) $\downarrow$ & ($^\circ$) $\downarrow$ \\
\midrule
Retargeting ($q^{\mathrm{init}}$) & \hc{3}25          & \hc{4}8.8          & \hc{3}34          & \hc{3}29          & \hc{3}4.6          & \hc{3}38          & \hc{3}22           & \hc{3}6.8         & \hc{3}40          & \hc{3}17          & \hc{3}7.0          & \hc{3}45          & \hc{3}12          & \hc{3}8.5          & \hc{3}53 \\
\quad + contact refinement        & \hc{4}21          & \hc{3}8.3          & \hc{3}34          & \hc{4}23          & \hc{4}5.4          & \hc{4}44          & \hc{4}19          & \hc{4}7.5          & \hc{4}46          & \hc{3}17          & \hc{4}7.7          & \hc{4}50          & \hc{4}9           & \hc{3}8.5          & \hc{4}59 \\
\quad + physics-in-the-loop       & \hc{2}36          & \hc{2}6.8          & \hc{2}27          & \hc{2}87          & \hc{1}1.7 & \hc{1}12 & \hc{1}88 & \hc{1}1.5 & \hc{1}10 & \hc{2}87          & \hc{2}1.8          & \hc{2}12          & \hc{1}85          & \hc{1}2.4          & \hc{1}14 \\
\quad + both                      & \hc{1}39 & \hc{1}6.7 & \hc{1}26 & \hc{1}89 & \hc{2}1.8          & \hc{1}12 & \hc{2}86          & \hc{2}1.6          & \hc{2}11          & \hc{1}89 & \hc{1}1.5 & \hc{1}10 & \hc{1}85 & \hc{1}2.4 & \hc{1}14 \\
\bottomrule
\end{tabular}
\end{table}

\section{Conclusion}

We presented \method{}, a training-free pipeline that turns a monocular RGB video of hand-object interaction into an interaction-faithful trajectory on a dexterous hand.
Instead of absorbing errors into a learned policy, it enforces physical consistency in reconstruction, retargeting and physics-in-the-loop refinement,
correcting each error in the stage where it arises.
Experiments show lower hand pose error and hand-object intersection in reconstruction,
markedly higher success than prior motion-transfer methods on five hands with 6 to 22 DoF,
and end-to-end success on videos where prior video-to-robot pipelines fail.

\paragraph{Limitations.}
\method{} is currently limited by monocular perception accuracy, especially metric scale estimation.
Physics-in-the-loop refinement is also affected by errors in the video-derived object geometry and estimated physical properties,
and could benefit from more accurate geometry and dynamics estimation from video.
Real-world execution is open-loop, so incorporating feedback control is a natural direction for improving robustness.

\bibliography{refs}
\bibliographystyle{natbib-authoryear}

\clearpage
\appendix
\section*{Appendix}
\phantomsection\addcontentsline{toc}{section}{Appendix}
\setcounter{subsection}{0}
\renewcommand{\thesubsection}{\Alph{subsection}}
\renewcommand{\theHsubsection}{appendix.\Alph{subsection}}
\crefalias{subsection}{appendix}
\subsection{Reconstruction details}
\label{app:recon}

\Cref{tab:recon_params} lists the key settings of \cref{sec:method:recon}; $\ell$ is the index distal phalanx length.
For the single-image benchmark of \cref{sec:exp:recon}, the intrinsics come from the dataset and the MoGe-3 depth is used at its own metric scale, since a single frame allows neither frame alignment nor a hand-defined unit.

\begin{table}[h]
\centering
\caption{Key reconstruction settings.}
\label{tab:recon_params}
\small
\begin{tabular}{@{}l p{0.76\linewidth}@{}}
\toprule
Component & Setting \\
\midrule
Segmentation & SAM 3.1; objects are grounded by their category words and propagated through the clip \\
Depth and camera & MoGe-3 (ViT-G); focal length is the median estimate over three frames, principal point at the image center \\
Frame alignment & similarity to the first keyframe on table pixels within 400\,px of the objects, 20\% residual trimming \\
Object mesh & SAM 3D on the frame with at most 5\% occlusion and the largest visible mask, decimated to 40k faces \\
Object pose & FoundationPose from the object mask in every frame, 5 refinement iterations \\
Hand placement & 3 rounds of the similarity fit with 10\% residual trimming; $\kappa=2.41$, i.e.\ $w_i=0.3$ at $d_i=\ell/2$ \\
Refinement gate & only hands within 10\,mm of the object are refined \\
Schedule & 5 cycles of 10 Adam~\citep{kingma2015adam} steps for each of the two stages; crossing repair after every cycle \\
Image alignment & learning rate $2\times10^{-3}$ for $\vtheta$ and $2\times10^{-2}$ for $\vbeta$; $\lambda_{\mathrm{ord}}=30$, $s_{\mathrm{o}}=5$\,mm \\
Contact-penetration & learning rate $10^{-3}$ for $\vtheta$ and $\mathbf{T}_h$; $\lambda_c=66$, $\lambda_p=7\times10^{5}$, $d_0=0.001\ell$, $\sigma_c=\sigma_p=0.5\ell$ \\
Crossing repair & 2\,mm clearance; mean vertex displacement at most 15\,mm, 5\,mm per cycle \\
Keyframes & sampling ratio $r=0.2$ \\
\bottomrule
\end{tabular}
\end{table}

\subsection{Adaptive keyframe sampling}
\label{app:keyframes}

$A_t$ in \cref{eq:motionsignal} climbs steeply during dynamic motions and stalls whenever the hand moves smoothly or holds still,
crowding keyframes into reversals and transients, and spending almost none on steady stretches, maximizing action fidelity under a fixed compression ratio (\cref{fig:keyframes}).
Each hand accumulates its own $A_t$ so a stationary hand does not pay for the other's keyframes.

\begin{table}[t]
\centering
\caption{Keyframe sampling on XHand, 100 tasks, ATE criterion. At each ratio $r$ both rules select the same number of keyframes and keep the first and last frame. Rollouts: number of physics simulations, i.e., candidate actions the optimizer tries out in simulation, which measures the actual compute (median over tasks).}
\label{tab:ablation_keyframes}
\scriptsize
\setlength{\tabcolsep}{2pt}
\begin{tabular}{c cccc cccc}
\toprule
 & \multicolumn{4}{c}{Adaptive} & \multicolumn{4}{c}{Uniform} \\
\cmidrule(lr){2-5} \cmidrule(lr){6-9}
$r$ & Succ. & Trans. & Rot. & Rollouts & Succ. & Trans. & Rot. & Rollouts \\
 & (\%) $\uparrow$ & (cm) $\downarrow$ & ($^\circ$) $\downarrow$ &  & (\%) $\uparrow$ & (cm) $\downarrow$ & ($^\circ$) $\downarrow$ &  \\
\midrule
0.05 & \textbf{57} & \textbf{4.4} & \textbf{29} & 0.9K & 50 & 5.6 & 33 & 1.0K\\
0.1 & \textbf{84} & \textbf{1.9} & \textbf{13} & 1.9K & 82 & 2.3 & 14 &  1.7K\\
0.2 & \textbf{83} & \textbf{1.9} & \textbf{12} & 3.4K & 81 & \textbf{1.9} & 13 & 3.4K \\
1 & -- & -- & -- & -- & \textbf{89} & \textbf{1.7}  & \textbf{12} & \textbf{13K}\\
\bottomrule
\end{tabular}
\end{table}

Physics-in-the-loop refinement runs one CMA-ES search per keyframe, so the sampling ratio $r$ sets its cost, measured in rollouts (\cref{tab:ablation_keyframes}).
On XHand we compare adaptive sampling with uniform sampling of the same keyframe budget over 100 tasks at $r\in\{0.05,0.1,0.2,1\}$, on the protocol of \cref{sec:exp:transfer} (\cref{tab:ablation_keyframes}).
At $r = 1$ the two rules coincide. At every lower ratio, adaptive sampling matches or exceeds uniform sampling on all three metrics with a comparable number of rollouts.
Rollouts do not fall in proportion to $r$: each keyframe needs 30--50\% more of them than at $r = 1$. With dense keyframes, the solution at one keyframe is nearly optimal for the next,
so the warm-started search meets its stopping criterion quickly. As keyframes grow sparser, consecutive keyframes differ more,
the previous solution becomes a poorer starting point, and each search needs more candidates to converge.

\subsection{Algorithms}
\label{app:algorithms}

\Cref{alg:crossing} gives the finger-crossing repair of \cref{sec:method:recon}, and \cref{alg:retarget} the contact-aware retargeting refinement of \cref{sec:method:exec}.

\begin{algorithm}[t]
\caption{Finger-crossing repair of a MANO hand.}
\label{alg:crossing}
\small
\begin{algorithmic}[1]
\Require MANO hand $(\vtheta_0, \vbeta, \mathbf{T}_{h,0})$, as placed or after a refinement cycle
\Statex $n_f$: number of points at which finger $f$'s keypoint polyline $\gamma_f$ (wrist--knuckle--middle--distal--tip)
\Statex crosses the object surface; odd: the tip is inside, nonzero even: the finger passed through
\Statex $F$: fingers to repair
\Statex $M$: metric on pose changes; $\|\Delta\|_M^2$ is the squared pixel motion of all MANO vertices in the image
\Statex when the pose changes by $\Delta$
\State $F \gets \{f : n_f > 0\}$
\State $(\vtheta, \mathbf{T}_h) \gets (\vtheta_0, \mathbf{T}_{h,0})$ \Comment{$\vbeta$ fixed; joints of $F$ and hand placement vary}
\While{$F \neq \emptyset$}
  \State $\rho_f \gets L_f$ if $n_f > 0$, else $d_f$ \Comment{$\rho_f$: residual of finger $f$}
  \Statex \Comment{$L_f$: length of $\gamma_f$ from its first entry point to the tip}
  \Statex \Comment{$d_f$: depth of $f$'s deepest vertex inside the object}
  \State $(\vtheta^\star, \mathbf{T}_h^\star) \gets$ the pose nearest to $(\vtheta_0, \mathbf{T}_{h,0})$ under $M$ at which every linearized $\rho_f$ vanishes
  \State move $(\vtheta, \mathbf{T}_h)$ toward $(\vtheta^\star, \mathbf{T}_h^\star)$ by the largest $\alpha \le 1$ that lowers $\sum_{f \in F} \rho_f$,
  \State $F \gets \{f \in F : n_f > 0$ or $f$ has vertices inside the object$\}$
\EndWhile
\State \Return $(\vtheta, \mathbf{T}_h)$, or $(\vtheta_0, \mathbf{T}_{h,0})$ if the repair fails or moves the hand too far
\end{algorithmic}
\end{algorithm}

\begin{algorithm}[t]
\caption{Contact-aware retargeting refinement.}
\label{alg:retarget}
\small
\begin{algorithmic}[1]
\Require solver pose $q^{\mathrm{init}}$; object samples $O$; vertices $V_p$, SDF $\psi_p$, anchors $A_p$ and region $R_p$ of $p$
\Statex $M$: metric on changes $\Delta q$ of joints and wrist; $\|\Delta q\|_M^2$ is the weighted squared motion of the anchors,
\Statex the wrist and the joints, with the joints weighted least
\Statex $\rho_p$: residual of $p$; $\rho = \big[\rho_{\mathrm{thumb}}^\top, \rho_{\mathrm{index}}^\top, \dots, \rho_{\mathrm{pinky}}^\top, \rho_{\mathrm{palm}}^\top\big]^\top$; $J_\rho = \partial \rho / \partial q$
\Statex $S = (N_{\mathrm{cross}}, \|\rho\|_1)$: the number of fingers with $n_p > 0$, then the summed residual, compared
\Statex lexicographically ($<_{\mathrm{lex}}$): fewer crossing fingers always ranks first
\State $q \gets q^{\mathrm{init}}$
\Repeat
  \For{$p \in$ \{fingers, palm\}}
    \If{$n_p > 0$} \Comment{$p$ crosses the object: pull it out first (\cref{alg:crossing})}
      \State $\rho_p \gets L_p$
    \Else \Comment{$p$ is outside: remove penetration and enclosure,}
    \Statex \CommentCont{\phantom{$p$ is outside: }reach the contact region}
      \State $\rho_p \gets \big(\operatorname{ReLU}(-\phi(V_p)),\ \operatorname{ReLU}(-\psi_p(O)),\ \operatorname{distance}(A_p, R_p)\big)$
    \EndIf
  \EndFor
  \State solve $(J_\rho^\top J_\rho + \mu M)\,\Delta q = -J_\rho^\top \rho$ \Comment{damping $\mu > 0$ keeps $\|\Delta q\|_M$ small}
  \State $q \gets q + \alpha\Delta q$ with the largest $\alpha \le 1$ that lowers $S$
\Until{no violation remains or no step lowers $S$}
\State \Return $q$ if $S(q) <_{\mathrm{lex}} S(q^{\mathrm{init}})$, else $q^{\mathrm{init}}$
\end{algorithmic}
\end{algorithm}

\subsection{Simulation settings}
\label{app:sim}

All methods are replayed in the same MuJoCo scene and scored on the same ground truth; only the hand trajectory differs.
\Cref{tab:sim_params} lists the shared settings.
Our method and OmniRetarget enter as keyframes interpolated with PCHIP (rotations with an $\mathrm{SO}(3)$ spline) on a 60\,Hz grid,
SPIDER as its commanded controls resampled from 100\,Hz to 60\,Hz (holding the last state if its rollout ends early),
and ManipTrans as its executed states, frame for frame.
Every method's commands are projected onto the joint coupling of the underactuated hands before replay.

\begin{table}[h]
\centering
\caption{Simulation settings shared by all methods in the transfer benchmark.}
\label{tab:sim_params}
\small
\begin{tabular}{@{}l p{0.74\linewidth}@{}}
\toprule
Component & Setting \\
\midrule
Solver & timestep 1/240\,s (4 steps per 60\,Hz frame); implicit-fast integrator; Newton solver with 10 iterations and 10 no-slip iterations; elliptic friction cones, impedance ratio 10 \\
Gravity & 9.81\,m/s$^2$ along the negative normal of the support plane \\
Contacts & condim 6; friction (1.0, 0.01, 0.0002); solref (0.003, 1), solimp (0.9, 0.99, 0.001); margin 1\,mm; the hands do not collide with the table \\
Table & plane taken from the ground-truth scene \\
Objects & ground-truth initial pose and velocity; cataloged masses (typical weights for TACO and OakInk2, manufacturer weights for ARCTIC) spread uniformly over CoACD~\citep{wei2022coacd} hulls (2\,mm, at most 32); objects that neither move nor are touched in the ground truth are fixed; every annotated object is present \\
Articulations & ARCTIC objects: limited hinge over the recorded range, damping 0.02\,N\,m\,s, no friction loss \\
Hands & official kinematic models with their joint damping, friction loss and torque limits; position servos with $k_p = 500$, $k_v = 3$ and armature 0.001 for every hand; collision from CoACD hulls of the visual meshes (0.5\,mm for Shadow, 2\,mm otherwise); fingertip--fingertip and fingertip--palm self-collision \\
Wrist & free root welded to a mocap target, solref (0.0083, 1), solimp (0.99, 0.999, 0.001) \\
Control & targets interpolated with cubic splines (rotation spline for the root) to every step; the whole clip is replayed open-loop from its first frame \\
\bottomrule
\end{tabular}
\end{table}

\subsection{Transfer ablation details}
\label{app:transfer_ablation}

Contact refinement's open-loop losses in \cref{tab:ablation_transfer} are mostly false successes: in the tasks where fingers pass fully through the object, the interlocked object follows the hand as it would in a real grasp and passes the ATE criterion without being held; the remaining losses come from contact lost between interpolated keyframes. After physics-in-the-loop refinement, contact refinement also lowers the fraction of frames with hand-object contact from 41.4\% to 35.1\%, while object errors fall on the Inspire Hand and SharpaWave and change by at most 0.1\,cm and 1$^\circ$ on the other three hands. The contacts it removes are thus largely interlocking caused by penetration rather than contacts that grasp or support the object.

\subsection{End-to-end failure analysis}
\label{app:e2e_failures}

Both baselines estimate the camera with MoGe and replay the hand at its reconstructed position.
TACO is filmed with a 24$^\circ$ telephoto lens, which MoGe reads as 42--52$^\circ$, so the baselines place the whole scene at about half its true distance: although the hand pose itself is accurate, the wrist lands 0.73--0.93\,m from its true position and the object moves by more than 1\,cm in only 2 of 101 replays.
Upstream image-based geometry estimation is thus the largest bottleneck of both baselines.
OakInk2's 70$^\circ$ head-mounted camera lies within MoGe's prior and the wrist error falls to 37--59\,mm, yet the hand pushes the object aside instead of grasping it, reproducing at most 3\% of the ground-truth contacts, so grasping is a second bottleneck, independent of depth.
\method{} fails on 28 clips, 14 in perception, 13 in execution and 1 in motion planning.
The perception failures reconstruct the object at the wrong size from a mask that misses or overruns it (12 clips), at the wrong shape (1), or fail to ground it (1); in the execution failures the hand lets the object rotate (7), never closes on it (5), or drifts off the ground-truth path (1).

\subsection{Real-world deployment}
\label{app:realworld}

\Cref{fig:realworld_strips_a,fig:realworld_strips_b} show all seven real-world tasks of \cref{sec:exp:real} as film strips.
For each task, the top row is the input human video and the bottom row the execution on the two XHands;
each row has seven frames evenly spaced in time over the motion in its video.
The two videos of a task differ in length and are sampled independently.

\makeatletter\setlength{\@fptop}{0pt}\setlength{\@fpsep}{\floatsep}\makeatother
\newcommand{\realstriprow}[2]{%
  \sbox0{#2}%
  \noindent\makebox[10pt][l]{\rotatebox{90}{\makebox[\ht0][c]{\scriptsize #1}}}\usebox0\par}

\begin{figure}[!h]
\noindent\hspace{10pt}{\small\textbf{insertion}}\par\vspace{2pt}
\realstriprow{Human}{\includegraphics[width=\dimexpr\linewidth-10pt\relax]{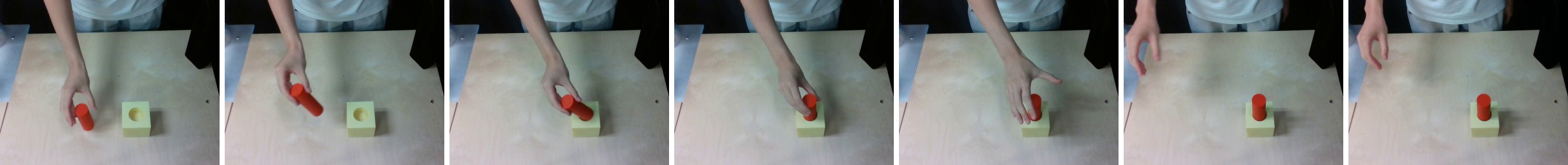}}\vspace{2pt}
\realstriprow{Robot}{\includegraphics[width=\dimexpr\linewidth-10pt\relax]{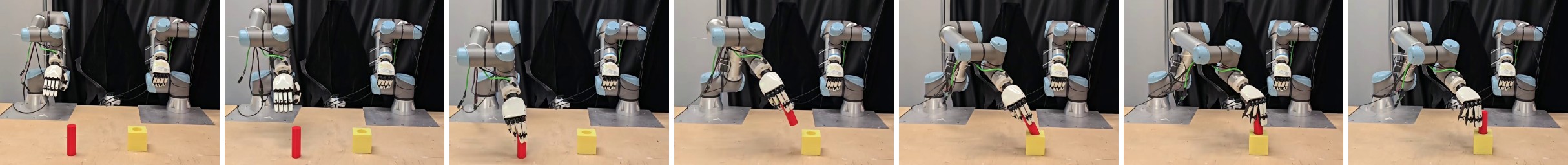}}\vspace{9pt}
\noindent\hspace{10pt}{\small\textbf{stacking}}\par\vspace{2pt}
\realstriprow{Human}{\includegraphics[width=\dimexpr\linewidth-10pt\relax]{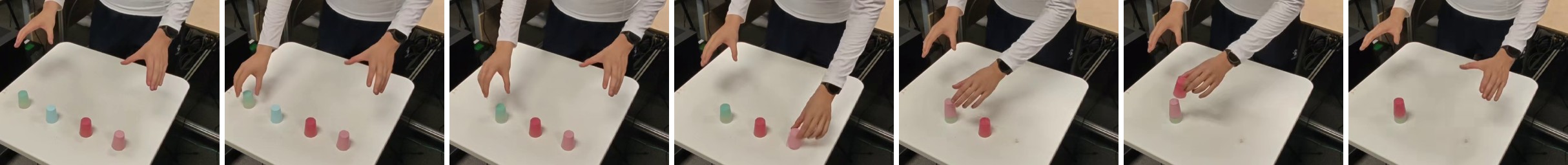}}\vspace{2pt}
\realstriprow{Robot}{\includegraphics[width=\dimexpr\linewidth-10pt\relax]{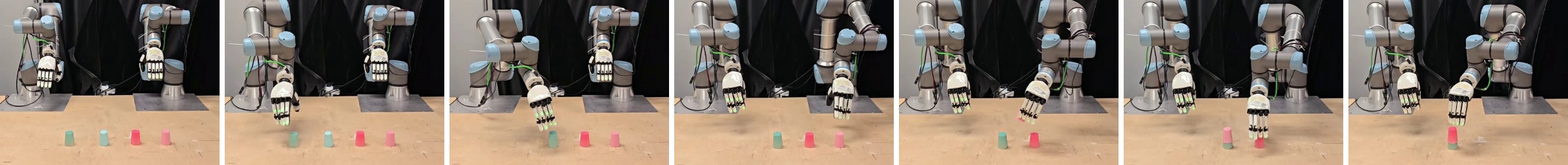}}\vspace{9pt}
\noindent\hspace{10pt}{\small\textbf{sweeping}}\par\vspace{2pt}
\realstriprow{Human}{\includegraphics[width=\dimexpr\linewidth-10pt\relax]{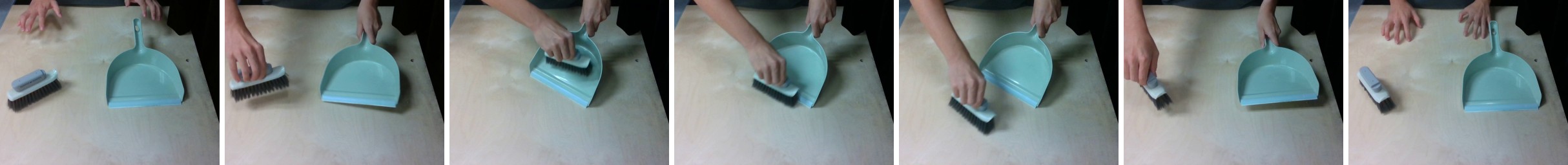}}\vspace{2pt}
\realstriprow{Robot}{\includegraphics[width=\dimexpr\linewidth-10pt\relax]{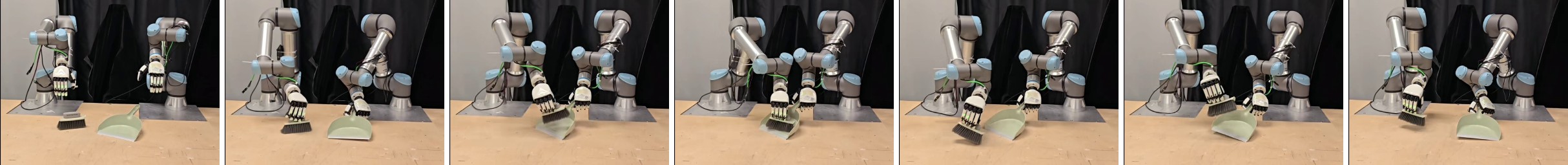}}\vspace{9pt}
\noindent\hspace{10pt}{\small\textbf{scanning}}\par\vspace{2pt}
\realstriprow{Human}{\includegraphics[width=\dimexpr\linewidth-10pt\relax]{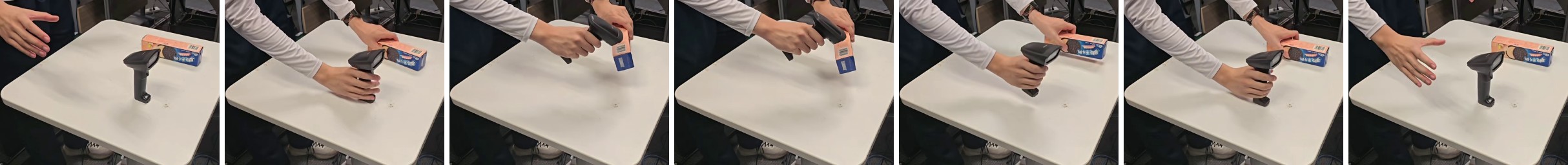}}\vspace{2pt}
\realstriprow{Robot}{\includegraphics[width=\dimexpr\linewidth-10pt\relax]{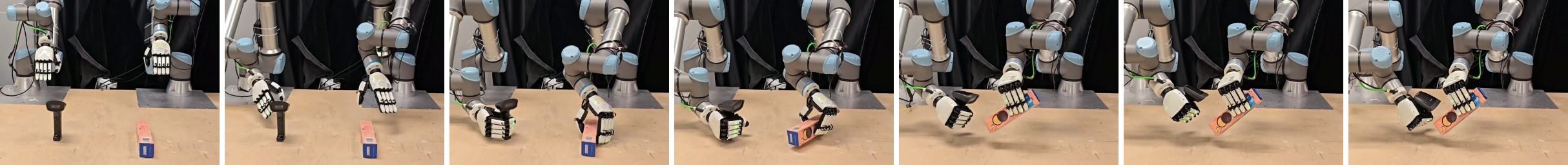}}\vspace{9pt}
\vspace{-6pt}
\caption{\textbf{Real-world film strips}: insertion, stacking, sweeping and scanning.
For each task, top: the input human video; bottom: the robot executing the trajectory.
Time runs from left to right.}
\label{fig:realworld_strips_a}
\end{figure}

\begin{figure}[!t]
\noindent\hspace{10pt}{\small\textbf{scrubbing}}\par\vspace{2pt}
\realstriprow{Human}{\includegraphics[width=\dimexpr\linewidth-10pt\relax]{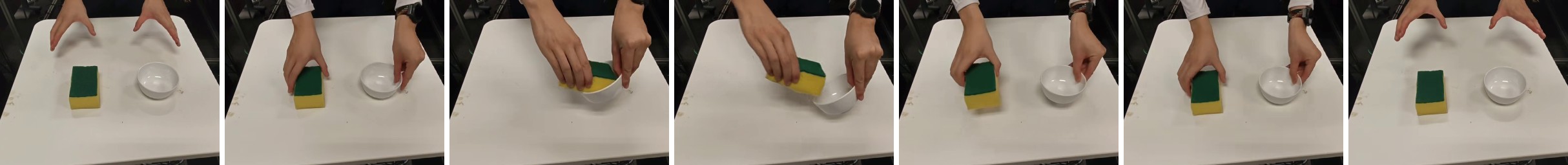}}\vspace{2pt}
\realstriprow{Robot}{\includegraphics[width=\dimexpr\linewidth-10pt\relax]{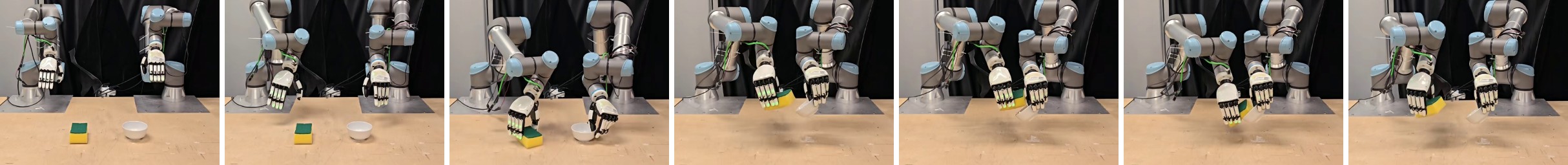}}\vspace{9pt}
\noindent\hspace{10pt}{\small\textbf{pouring (mug)}}\par\vspace{2pt}
\realstriprow{Human}{\includegraphics[width=\dimexpr\linewidth-10pt\relax]{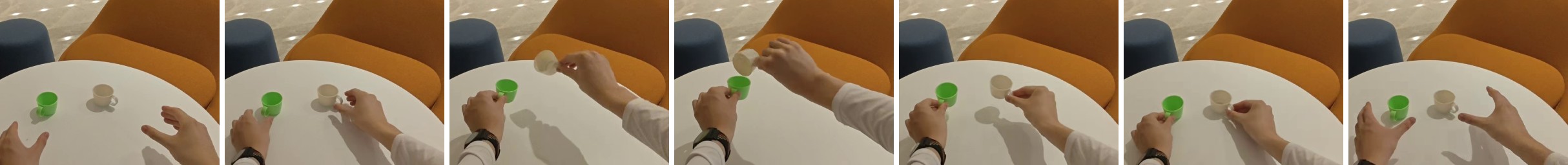}}\vspace{2pt}
\realstriprow{Robot}{\includegraphics[width=\dimexpr\linewidth-10pt\relax]{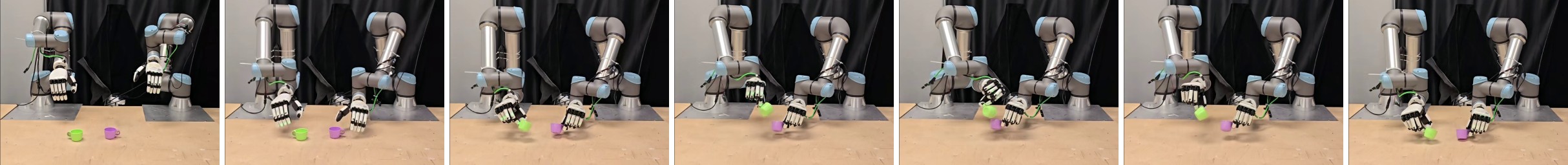}}\vspace{9pt}
\noindent\hspace{10pt}{\small\textbf{pouring (cup)}}\par\vspace{2pt}
\realstriprow{Human}{\includegraphics[width=\dimexpr\linewidth-10pt\relax]{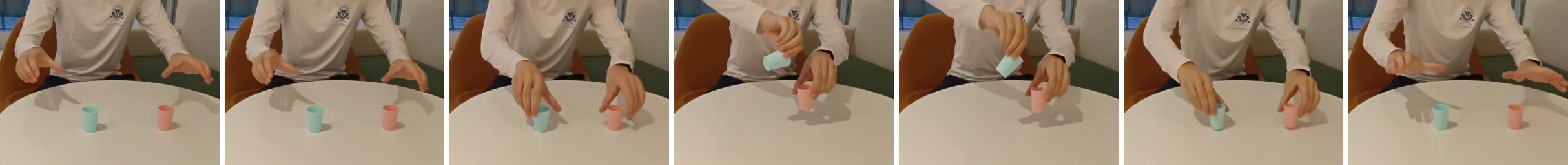}}\vspace{2pt}
\realstriprow{Robot}{\includegraphics[width=\dimexpr\linewidth-10pt\relax]{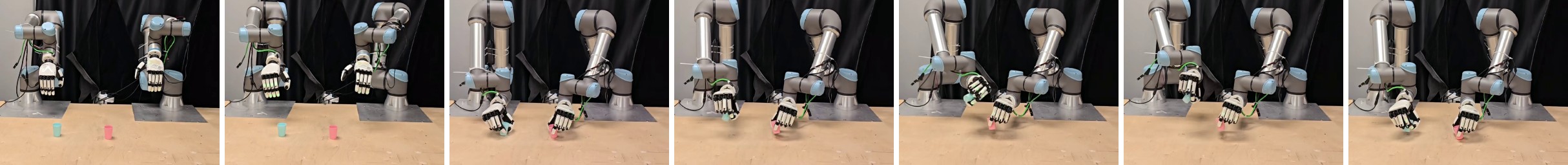}}\vspace{9pt}
\vspace{-6pt}
\caption{\textbf{Real-world film strips}, continued: scrubbing and the two pouring tasks, laid out as in \cref{fig:realworld_strips_a}.}
\label{fig:realworld_strips_b}
\end{figure}

\end{document}